\documentclass[letterpaper]{article} 
\usepackage{aaai2026}  
\nocopyright
\usepackage{times}  
\usepackage{helvet}  
\usepackage{courier}  
\usepackage[hyphens]{url}  
\usepackage{graphicx} 
\usepackage{natbib}  
\usepackage{caption} 
\usepackage{amssymb}
\usepackage{algorithm}
\usepackage{algorithmic}
\usepackage{booktabs}
\usepackage{subcaption}
\usepackage{amsmath}
\usepackage{pgfmath}
\usepackage{newfloat}
\usepackage{listings}
\DeclareCaptionStyle{ruled}{labelfont=normalfont,labelsep=colon,strut=off} 
\floatstyle{ruled}
\newfloat{listing}{tb}{lst}{}
\floatname{listing}{Listing}
\title{EMMM, Explain Me My Model! Explainable Machine Generated Text Detection in Dialogues}
\author {
    \equalcontrib Angela Yifei Yuan\textsuperscript{\rm 1},
    \equalcontrib Haoyi Li\textsuperscript{\rm 1},
    Soyeon Caren Han\textsuperscript{\rm 1},
    Christopher Leckie\thanks{Corresponding author}\textsuperscript{\rm 1}
}
\affiliations {
    \textsuperscript{\rm 1}The University of Melbourne\\
    \{angela.yuan, haoyil4\}@student.unimelb.edu.au,
    \{caren.han, caleckie\}@unimelb.edu.au
}

\usepackage{bibentry}

\begin{document}

\maketitle

\begin{abstract}

The rapid adoption of large language models (LLMs) in customer service introduces new risks, as malicious actors can exploit them to conduct large-scale user impersonation through machine-generated text (MGT). Current MGT detection methods often struggle in online conversational settings, reducing the reliability and interpretability essential for trustworthy AI deployment.
In customer service scenarios where operators are typically non-expert users, explanation become crucial for trustworthy MGT detection. In this paper, we propose EMMM, an explanation-then-detection framework that balances latency, accuracy, and non-expert-oriented interpretability. Experimental results demonstrate that EMMM provides explanations accessible to non-expert users, with 70\% of human evaluators preferring its outputs, while achieving competitive accuracy compared to state-of-the-art models and maintaining low latency, generating outputs within 1 second. Our code and dataset are open-sourced\footnote{https://github.com/AngieYYF/EMMM-explainable-chatbot-detection}.
\end{abstract}

\section{Introduction} \label{intro}
Large language models (LLMs) have revolutionized human–AI interaction, enabling highly realistic conversations across a wide range of applications. However, this progress also brings serious security risks: malicious actors can exploit LLMs to impersonate users and launch large-scale attacks on online platforms. Such misuse can disrupt critical services such as emergency response and customer support, leading to denial-of-service incidents and operational failures~\cite{Owasp2025}. As LLMs continue to proliferate, reliable and explainable detection of machine-generated text (MGT) in conversational settings has become essential for safeguarding platform reliability and user trust.
Existing detectors focus on isolated text passages, and they often struggle to adapt to the dynamic and interleaved structure of real-world conversations. Moreover, trust in MGT detection systems, especially in online conversational environments, requires explanations that are interpretable to diverse users, including non-experts. In such settings, explanations must be clear and accessible to support understanding of model behavior, review flagged cases, and inform moderation decisions. To address this, we propose EMMM, a framework specifically designed for real-time MGT detection in conversational settings. By leveraging conversation structure, EMMM achieves high detection accuracy while offering multi-level, multi-dimension, and multi-strategy explanations tailored for broad user accessibility.
We identify three key challenges unique to explainable MGT detection in conversational settings. First, asymmetric detection in dialogues, where interactions are interleaved but detection targets only one party, creates an unusual input structure that limits the model from using full conversational context and necessitates specialized handling.
Second, user-friendly explanations for non-experts remain limited, as current methods frequently use technical metrics such as feature weights~\cite{Shah2023-XAIGTDetection, Schoenegger2024-XAIGT} which are inaccessible to service operators without technical backgrounds. 
Third, local attribution explanations in MGT detection models are difficult to interpret due to the absence of ground-truth, whereas globally aggregated explanations lack the granularity needed for instance-level interpretation.

\begin{figure*}[t]
    \centering
    \includegraphics[width=\linewidth]{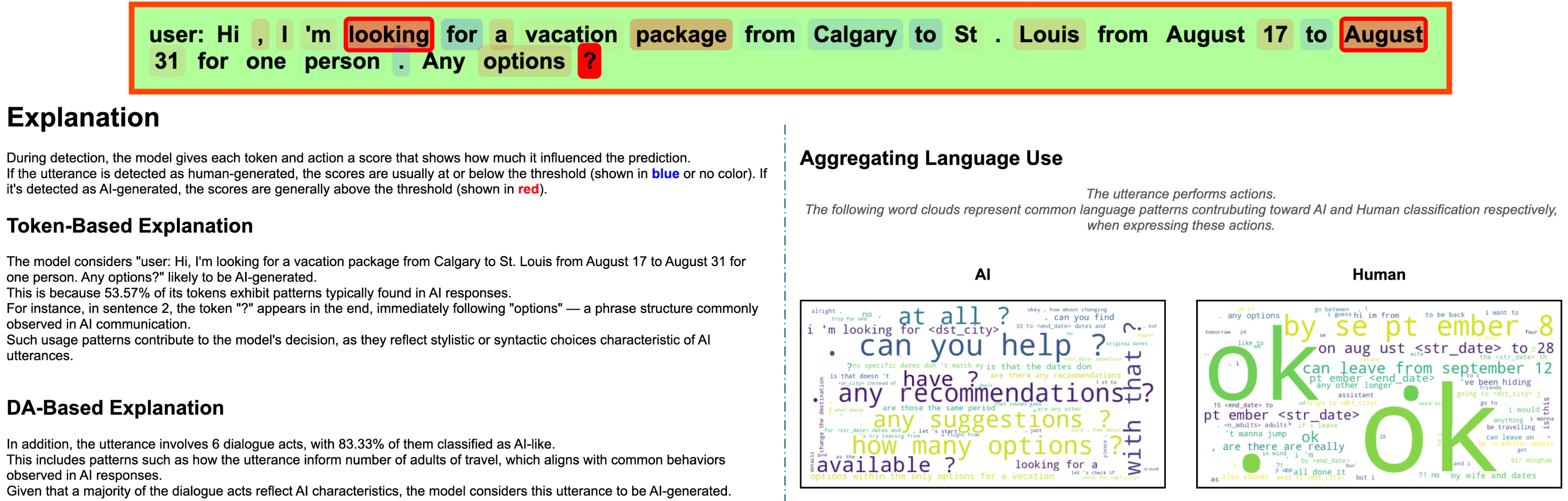}
    \caption{A demonstration of EMMM framework online detection and non-expert oriented explanation.}
    \label{fig:demo-explanation}
\end{figure*}

To address these challenges, we propose EMMM, an explanation-driven framework for interpretable LLM chatbot detection in conversational settings. EMMM integrates (1) \textbf{Multi-dimensional} inputs (behaviors and language), operates at (2) \textbf{Multi-level} interaction (turn and dialogue), and employs (3) \textbf{Multi-strategy} explanation (local natural language explanations, and semi-global visual insights). Guided by speech act theory~\cite{Austin1975-speechActTheory}, EMMM explicitly incorporates dialogue acts into its design. It processes each incoming user utterance through turn-level and dialogue-level detection modules, selectively aggregating important features across turns. The system generates an explanation report that combines highlighted features, natural language reasoning, and semi-global visualizations, as illustrated in Figure~\ref{fig:demo-explanation}.
This approach effectively addresses asymmetric detection by isolating and selectively processing the target party’s utterances within dialogue turns. The generated explanation report enhances accessibility for non-expert users by articulating raw attribution data in natural language, and improves MGT detection interpretability by incorporating semi-global model insights to balance local relevance with global interpretability.

Contributions of EMMM can be summarized as follows:
\begin{itemize}
\item \textbf{EMMM is Dialogue-Aware.}
EMMM leverages conversation specific features to deliver multi-dimension, multi-level, and multi-strategy explanations.  Grounded in speech act theory, it models dialogue structure and intent to enhance interpretability. EMMM supports both online and offline chatbot detection, achieving a balance between detection performance and explanation quality.

\item \textbf{EMMM is Efficient.}
EMMM produces explanation reports online in under 1 second by combining a sequential selector–predictor pipeline with offline preprocessing, achieving the time efficiency required for deployment in real-world service platforms.

\item \textbf{EMMM is Interpretable.}
EMMM generates non-expert user friendly natural language explanation reports and includes visualizations of contextualized semi-global model behaviors to enhance model interpretability. We evaluate interpretability via qualitative analysis and a human survey, showing strong user preference of $69\%$ over a baseline attribution approach.
\end{itemize}

To the best of our knowledge, this is the first framework to tackle the challenging problem of explainable MGT detection for non-expert users in conversational settings, paving the way for practical, human-aligned AI safety solutions.

\section{Related Work}

\begin{figure*}[t]
    \centering
    \includegraphics[width=0.8\linewidth]{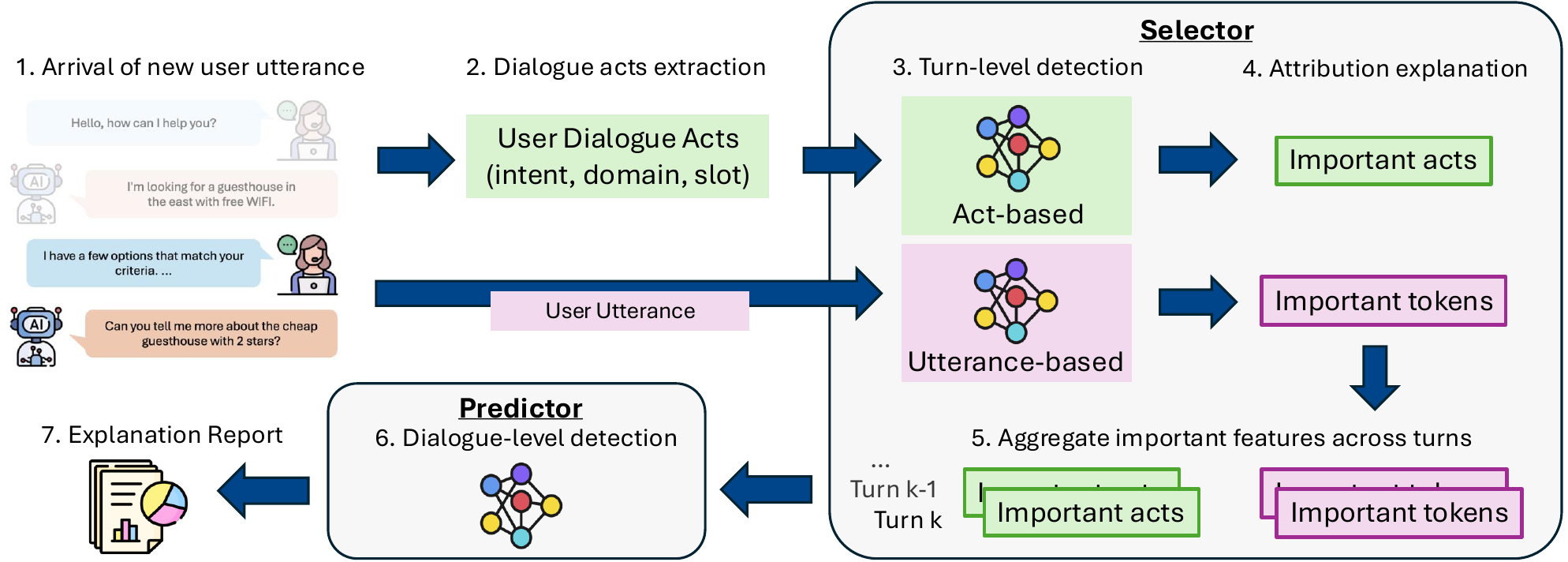}
    \caption{The 7 steps process of the EMMM explainable detection framework.}
    \label{fig:framework}
\end{figure*}

\subsection{MGT Detection}
A wide range of methods have been explored for MGT detection, which aims to classify whether a given text was produced by a human or by a LLM. Approaches include watermarking techniques that embed hidden signals in generated content~\cite{Lu2024-watermark, Kirchenbauer2024-watermark}, as well as zero-shot and supervised classification models. 
For example, Binoculars~\cite{Hans2024-binoculars} is a statistical zero-shot method using two language models to compute an AI-likelihood score based on entropy differences, requiring no training data. In contrast, supervised models excel in specialized tasks by learning meaningful patterns from labeled training data~\cite{Wu2025-AIGTSurvey, Bafna2024-AIGTroberta}.

Existing work on MGT detection primarily focuses on text passages, overlooking the complexity and interpretative richness of online dialogues.
To address this, we propose EMMM tailored for real-time detection of LLM chatbots in online dialogues, leveraging dialogue-specific features to provide explanations that non-expert users can interpret.

\subsection{Explaining MGT Detection}
Explaining a MGT detection model involves understanding its decision-making process to foster user trust and evaluate model performance ~\cite{Luo2024-XNLPSurvey}.
There are two common local explanation approaches, attribution-based explanations and Natural Language Explanations (NLE).

\subsubsection{Attribution-based Explanations}
Attribution methods assign importance scores to input features for a specific prediction~\cite{Sundararajan2017-integratedGradient, Tsai2023-faithShap}. This is the primary method used in existing MGT detection work, identifying importance of extracted attributes in feature-based detection \cite{Shah2023-XAIGTDetection}, or tokens and phrases in pretrained language model (PLM)–based approaches \cite{Schoenegger2024-XAIGT}.
Token and phrase attribution has been widely applied in tasks such as sentiment analysis and reading comprehension, where the importance scores are expected to align with intuitively relevant input regions for interpretability~\cite{Tsang2020-binUnion}. 
However, MGT detection lacks ground-truth important tokens, making interpretation more challenging. 
Global aggregation of local explanations can reveal broader model patterns to assist interpretation~\cite{Mor2024-aggregation} but lacks relevance to individual predictions. 
We address this by providing semi-global model insights contextualized by target samples, enabling a higher-level yet relevant understanding of model behavior.

\subsubsection{Natural Language Explanations (NLE)}
NLE methods generate textual explanations for model predictions. Unlike raw attribution scores, which can be difficult for users to interpret, they aim to produce explanations that better align with human understanding and values. These methods fall into two categories: generation-based and template-based.
In generation-based approaches, the explanation models produce the entirety of the explanation content~\cite{Luo2024-XNLPSurvey}. While flexible, they often require datasets with labeled explanations for training~\cite{Yordanov2022-NLEtransferLearning, Marasovic2022-NLEfewShot}. This poses a challenge for MGT detection due to the lack of NLE-labeled data, and limited understanding of the differences between human and LLM chatbots in dialogue settings for NLE labeling. Prior attempts at NLE annotation targeted text passages and relied on domain experts, which is costly and not feasible at scale~\cite{Ji2024-XAIGTDetection, Russell2025-humanDetection}.
Template-based approaches offer a more practical alternative to enhance interpretability of raw explanation data, by defining explanation sentence templates filled in per sample~\cite{Zhang2020-templateNLE}. 
We propose a novel template using attribution insights to generate human-readable and interpretable explanations efficiently. This approach does not require prior intuition about the detection task for NLE annotation.

\subsection{Explanations for Non-experts}

Human-understandable explanations are widely recognized for their benefits, including improved efficiency and broader stakeholder coverage~\cite{CAMBRIA2023103111}. They aim to convey a model’s decision-making process to non-expert users, often through natural language~\cite{naturallangugeexplanclassificationburton2023} or intuitive visualizations~\cite{Kang2025AudioGenX} that make complex model behavior more accessible. Recent work has incorporated human–computer interaction (HCI) principles to guide the design of explanations, with naturalness, flexibility, and usefulness consistently emerging as critical factors~\cite{chromik2021XAIhumaninterface, Ji2024-XAIGTDetection}. Building on these insights, our research focuses on these three aspects in designing explanation reports, resulting in explanations that are better aligned with human values.

\section{Our Framework - EMMM} \label{sec:proposedApproach}
Due to the unique characteristics of multi-turn dialogues and the need for transparency in model decisions, we propose EMMM, an explainable detection framework designed to address key challenges in MGT detection within conversational settings. EMMM stands for Multi-Dimensional and Multi-Level detection and explanation, and Multi-Strategy explanation reporting. The overall process of EMMM is illustrated in Figure~\ref{fig:framework} and can be summarized as follows for each turn during an online conversation:
\begin{enumerate}
    \item \textbf{Arrival of new user utterance}: A new user utterance arrives, with sensitive information masked.
    \item \textbf{Dialogue acts extraction}: dialogue acts (DAs) are extracted from the user utterance. 
    \item \textbf{Turn-level detection}: Multi-dimensional detection is performed using the DAs and the utterance respectively.
    \item \textbf{Attribution explanation}: Important features (DAs and tokens) from the current turn are identified.
    \item \textbf{Aggregate important features across turns}: DAs and utterances from all turns are concatenated, while unimportant features are replaced by mask tokens.
    \item \textbf{Dialogue-level detection}: Detection is performed using the masked dialogue-level features.
    \item \textbf{Explanation report}: An explanation report is generated.
\end{enumerate}
During offline chatbot detection, all utterances are readily available to undergo steps 2 to 5 before the framework proceeds to step 6 for dialogue-level detection.

With the overall workflow established, the following sections describe the core components of EMMM, detailing how it realizes its objectives through: (i) Multi-Dimensional detection and explanation, integrating both linguistic signals and user behavior for richer interpretability; (ii) Multi-Level detection and explanation, enabling efficient hierarchical detection across turns and dialogues; and (iii) Multi-Strategy explanation reporting, delivering natural language local explanations complemented by semi-global visual insights for enhanced transparency.

\subsection{Multi-Dimensional Detection and Explanation}
Speech act theory states that language use not only conveys information but also performs actions through utterances~\cite{Austin1975-speechActTheory}. Dialogue Acts (DAs) encode the speaker’s intent and the pragmatic function of each utterance. For example, the DA \textit{(inform, hotel, area, west)} indicates a user informing a hotel area preference. Existing approaches to MGT detection and explanation predominantly rely on raw tokens, overlooking such structured communicative functions that offer a behavior-oriented perspective beyond surface-level linguistic cues. This motivates our use of DA extraction to capture user intent and support richer Multi-Dimensional explanations. We omit the value element (e.g., “\textit{west}”) to abstract away from specific slot values and focus on the underlying behavioral intent.

\subsection{Multi-Level Detection and Explanation}

Multi-turn dialogues arrive incrementally, and users expect real-time feedback. As feature attribution costs grow exponentially with accumulating features across turns, dialogue-level explanations become computationally expensive. To enable efficient Multi-Level explanation, EMMM employs a sequential selector–predictor design~\cite{Luo2024-XNLPSurvey}: the selector identifies important features from turn-level attributions, which are concatenated and passed as the dialogue-level explanation to the dialogue-level predictor. This eliminates the need to recompute the full explanation with each new utterance, and remains faithful by ensuring the predictor relies only on the provided explanations.

\subsection{Multi-Strategy Explanation Report}
Our proposed explanation reporting integrates two complementary strategies: local narrative explanations and contextualized semi-global visual insights. Unlike prior single-modality approaches, this combination unites natural language narratives with visual representations, yielding explanations that enhanced the understanding of non-expert users. An example explanation report is shown in Figure~\ref{fig:demo-explanation}.

\subsubsection{Narrative Explanation}

For narrative explanations, we design a lightweight natural language template to meet the requirement of low computational complexity. It includes a background introduction providing context about the input and classification task, as well token-level and DA-level explanations that use natural language to aid human interpretation of the attributions. 
Inspired by HCI principles for explainable AI~\cite{chromik2021XAIhumaninterface} and discourse analysis techniques~\cite{discourseanalysis}, we prioritize three criteria: naturalness, flexibility, and usefulness.
Naturalness captures how easily users can comprehend the model’s output. Flexibility reflects the ability to convey multiple perspectives and information types. Usefulness measures the effectiveness of the explanation to resolve users' potential confusion about detection results. 
Discourse-aware elements, such as highlighting rhetorical patterns, help bridge the gap between technical attributions and non-expert understanding.
The template was iteratively refined with input from experts and user feedback, to ensure clarity and accessibility without compromising efficiency.

\subsubsection{Contextualized Semi-global Aggregation}

\begin{algorithm}[t]
\caption{Semi-Global Aggregation - Offline}
\label{alg:offline-aggregation}
\textbf{Input}: Dataset $\mathcal{D}$ with utterances and Dialogue Acts (DAs)\\
\mbox{\textbf{Output}: Aggregated scores $A$, top features $F$}
\hspace*{\fill}
\par
\begin{algorithmic}[1]
\STATE $A, F \leftarrow \{\}, \{\}$ \hfill // Initialize scores and features
\FOR{each DA $\in$ DA\_types($\mathcal{D}$)}
    \STATE $T_{\text{DA}} \leftarrow$ 
    ExtractFeaturesPerDA($\mathcal{D}, \text{DA}$)
    \STATE $L \leftarrow$ GetLocalAttributionScores($T_{\text{DA}}$)
    \FOR{each class $c \in \{\text{AI}, \text{Human}\}$}
        \STATE $A[\text{DA}][c] \leftarrow$ GlobalAggregation($L, c$)
        \STATE $F[\text{DA}][c] \leftarrow$ TopK(A[DA][c])
    \ENDFOR
\ENDFOR
\RETURN $A$, $F$
\end{algorithmic}
\end{algorithm}

\begin{algorithm}[t]
\caption{Semi-Global Aggregation - Online}
\label{alg:online-aggregation}
\textbf{Input}: Target DAs $\mathcal{DA}_{\text{utt}}$, target class $c$, top features $F$\\
\mbox{\textbf{Output}: Semi-global important features $\mathcal{S}$}
\hspace*{\fill}
\par
\begin{algorithmic}[1]
\STATE $S \leftarrow \{\}$ \hfill // Initialize feature-score map
\FOR{each DA $\in \mathcal{DA}_{\text{utt}}$}
    \FOR{each $(f, s) \in F[\text{DA}][c]$}
        \STATE $S[f] \leftarrow S[f] + s$ \hfill // Accumulate score
    \ENDFOR
\ENDFOR
\STATE \textbf{return} $S$
\end{algorithmic}
\end{algorithm}

To complement the local narrative explanation, we propose Contextualized Semi-Global Visualization, which addresses the limited relevance of global model insights to specific target samples. Existing methods aggregate local explanations across datasets to derive global insights, but often fail to reflect the context of individual samples, limiting their support for instance-level user understanding. Our method leverages dialogue acts, grouping utterance sub-strings based on their conveyed DA and aggregating local explanations within each DA category. This contextualized aggregation produces semi-global insights that better align with the target sample.

Algorithms~\ref{alg:offline-aggregation} and~\ref{alg:online-aggregation} describe 
the computation of DA-based aggregation scores across a dataset and their use in the online explanation process for a target utterance.
A feature refers to a token or phrase extracted from text spans associated with a specific DA. 
During the offline phase (Algorithm~\ref{alg:offline-aggregation}), local attribution scores are computed for each token in the dataset using attribution methods, and phrase-level scores are obtained by averaging the attributions of constituent tokens. Features are grouped by DA type, and attribution scores are aggregated across features within the same group using global aggregation methods, producing DA-specific attribution profiles for each class. The top ranked features per DA for the AI and Human classes are recorded with their associated scores.
During online application (Algorithm~\ref{alg:online-aggregation}), semi-global important features are extracted by retrieving and accumulating scores of the top features for each DA in the target utterance. 
This DA-aware aggregation balances the interpretability of global model insights with the contextual relevance of local explanations. Implementation details, including feature matching between DAs and text spans, and word cloud construction, are provided in Appendix~\ref{appendix:DA-aggregation}.

\subsection{EMMM Implementation}\label{EMMM implementation}
EMMM supports flexible integration of different detection models and feature attribution methods. 
We conducted extensive experiments based on detection performance to determine an effective configuration. The chosen implementation fine-tunes a DistilGPT2 model~\cite{Sanh2019-DistilBERT} for both turn-level act-based and utterance-based detection. For feature attribution, we apply Faith-SHAP~\cite{Tsai2023-faithShap} to identify up to three dialogue acts and three tokens per utterance with the highest absolute attribution scores. The turn-level detection models are further fine-tuned using the most influential acts and tokens across all turns to enable dialogue-level detection. Act and token embeddings are combined via average fusion and passed into the final classification layer. DA extraction uses a supervised model~\cite{Zhu2023-ConvLab} for the SPADE dataset, and few shot prompting Qwen2.5-7B~\cite{qwen2.5} for Frames. Experimental details are provided in Appendix~\ref{appendix:experiment-details}.

\section{Experimental Methodology} \label{setup}
\subsection{Datasets}
We experiment on two datasets. SPADE~\cite{Li2025-SPADE} is the only benchmark for LLM chatbot detection in conversational settings, containing hotel-domain bona fide from the MultiWOZ dataset~\cite{Eric2019-MultiWOZ} and LLM-generated dialogues. We use its End-to-End Conversation dataset, where two LLM instances simulate a conversation, acting as system and user respectively to achieve the user goal~\cite{Li2025-SPADE}. 
To expand domain coverage, we extend its data generation framework to the travel-domain Frames dataset~\cite{Schulz2017-Frame} using Qwen2.5-32B~\cite{qwen2.5}.
This results in a new dataset comprising 1364 pairs of bona fide and synthetic dialogues. Dataset generation details are provided in  Appendix~\ref{appendix:dataset-construction}. We randomly divide each dataset by dialogue ID into training, validation, and test sets in a 70\%/15\%/15\% ratio.

\subsection{Evaluation Metrics} 
The framework is evaluated on four key criteria essential for human-aligned deployment of LLM chatbot detection models in real-world applications:
\begin{itemize}
    \item \textbf{Detection Performance} is measured by Macro-F1 score, where higher values indicate better performance. Supervised non-deterministic models are run four times, and results are averaged.
    \item \textbf{Explanation Relevance} is measured by $AOPC_k(G, c)$ metric~\cite{Mor2024-aggregation}, which quantifies the impact of the top $k$ features from aggregation $G$ on class $c$ predictions. Higher scores reflect stronger relevance and alignment with model behavior.
    \item \textbf{Interpretability} is evaluated via a human survey, comparing user preference between EMMM explanation report and the baseline attribution explanation.
    \item \textbf{Time Complexity} for each step of the framework is measured in seconds per utterance.
\end{itemize}

\subsection{Baselines}
For detection performance, we compare EMMM against a range of existing MGT detection models, including zero-shot, pretrained supervised, and fully trained supervised approaches. We use Binoculars~\cite{Hans2024-binoculars} as a state-of-the-art zero-shot baseline, following the optimal settings reported in its original paper. For pretrained supervised models, we evaluate RADAR~\cite{Hu2023-RADAR} and $\text{ChatGPTD}_{\text{roberta}}$~\cite{Guo2023-HC3} using their officially released weights without additional fine-tuning. We also compare against supervised models trained on the target datasets, including $\text{Raidar}_{\text{llama}}$~\cite{Mao2024-Raidar} using their llama2\_7b\_chat implementation, entropy-based detection~\cite{Gehrmann2019-statistical, Li2025-SPADE} computed with TF-IDF features, and both random forest and multilayer perceptron (MLP) trained on TF-IDF embeddings. Model and training details are in Appendix~\ref{appendix:experiment-details}.

For explanation interpretability evaluation, our explanation report is compared against the content of local feature attribution methods, which is the primary existing approach for explaining MGT detection models~\cite{Shah2023-XAIGTDetection, Schoenegger2024-XAIGT}.

\begin{figure*}[t]
    \centering
    \includegraphics[width=0.8\linewidth]{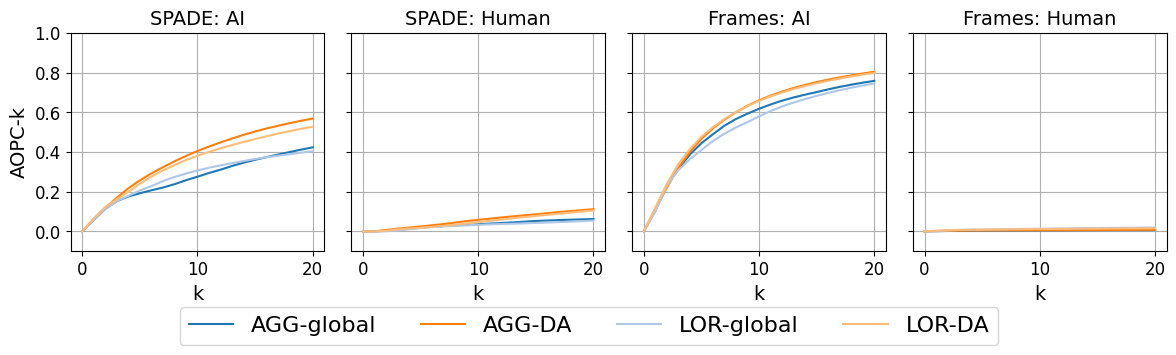}
    \caption{Comparison of explanation relevance between semi-global (\textit{-DA}) and global (\textit{-global}) aggregations using $AOPC_k(G, c)$ scores (y-axis) across different values of $k$ (x-axis). Semi-global aggregation  consistently outperforms global aggregation across all datasets and metrics (Wilcoxon signed\text{-}rank test, p $<$ 0.05).}
    \label{fig:aopc}
\end{figure*}

\section{Experimental Results}

\subsection{Detection Performance: Does EMMM detect MGT accurately?}

\begin{table}[t]
    \centering
    \begin{tabular}{lccc}
    \toprule
        \textbf{Detection Model} & \textbf{SPADE} & \textbf{Frames} & \textbf{Average} \\
        \hline
        \textit{Zero-shot} & & & \\
        \hline
        Binoculars\textsuperscript{1} &  0.5361 & 0.8115 & 0.6817 \\
        Binoculars\textsuperscript{2} & 0.8272  & 0.8191 & 0.8232 \\ 
        \hline
        \multicolumn{4}{l}{\textit{Pre-trained Supervised}} \\
        \hline
        $\text{ChatGPTD}_{\text{roberta}}$ & 0.5787 & 0.3989 & 0.4888 \\
        RADAR & 0.3452 & 0.4738 & 0.4094 \\ 
        \hline
        \multicolumn{4}{l}{\textit{Fully Supervised}} \\
        \hline
        Entropy &  0.5990 & 0.6539 & 0.5572\\
        $\text{Raidar}_{\text{llama}}$ & 0.7471 & 0.7945 & 0.7708 \\
        Random Forest & 0.9733 & 0.9902 & 0.9818 \\
        MLP & 0.9906 & 0.9976 & 0.9941 \\
        \hline
        \textbf{EMMM (ours)} & \textbf{0.9771} & \textbf{0.9945} & \textbf{0.9858} \\
    \bottomrule
    \end{tabular}
    \caption{Comparison of offline detection Macro-F1. Binoculars\textsuperscript{1} uses the default threshold of 0.9015, whereas Binoculars\textsuperscript{2} uses tuned thresholds based on a validation set (0.7777 for SPADE and 0.9038 for Frames).}
    \label{tab:baseline-performance}
\end{table}

During detection, user utterances are extracted from the dialogues and concatenated as input to the detection models. Table~\ref{tab:baseline-performance} presents the Macro-F1 score of detection models. 
Our framework, EMMM, maintains state-of-the-art detection performance, achieving an average Macro-F1 of 0.9858. As detailed in Section~\ref{EMMM implementation}, EMMM employs a sequential selector–predictor pipeline designed to provide efficient and detailed interpretability, avoiding the costly generation of explanations for the entire dialogue. Despite using only three tokens and three dialogue acts per utterance, EMMM consistently outperforms zero-shot, pre-trained supervised, and most fully supervised baselines. 
This shows that our proposed framework offers reliable detection performance while advancing the efficiency and explainability required for real-world, human-aligned deployment.

\begin{table}[t]
    \centering
    \begin{tabular}{lcc}
    \toprule
        \textbf{Attribution} & \textbf{1DA + 1token} & \textbf{3DA + 3token} \\
        \hline
        \textit{SPADE} & & \\
        \hline
        Faith-SHAP & \textbf{0.9449} & \underline{0.9771} \\
        STII & 0.9167 & 0.9233 \\
        Integrated gradient & \underline{0.9448} & \textbf{0.9866} \\
        \hline
        
        \textit{Frames} & & \\
        \hline
        Faith-SHAP & \underline{0.9835} & \textbf{0.9945} \\
        STII & \textbf{0.9841} & \underline{0.9939} \\
        Integrated gradient & 0.9774 & \underline{0.9939} \\
        
    \bottomrule
    \end{tabular}
    \caption{EMMM detection performance (Macro-F1) across attribution methods under varying interpretability constraints (number of features per utterance). Best scores per group are \textbf{bolded}, second-best are \underline{underlined}.}
    \label{tab:attribution-test}
\end{table}

\begin{figure}[t]
    \centering
    \includegraphics[width=\linewidth]{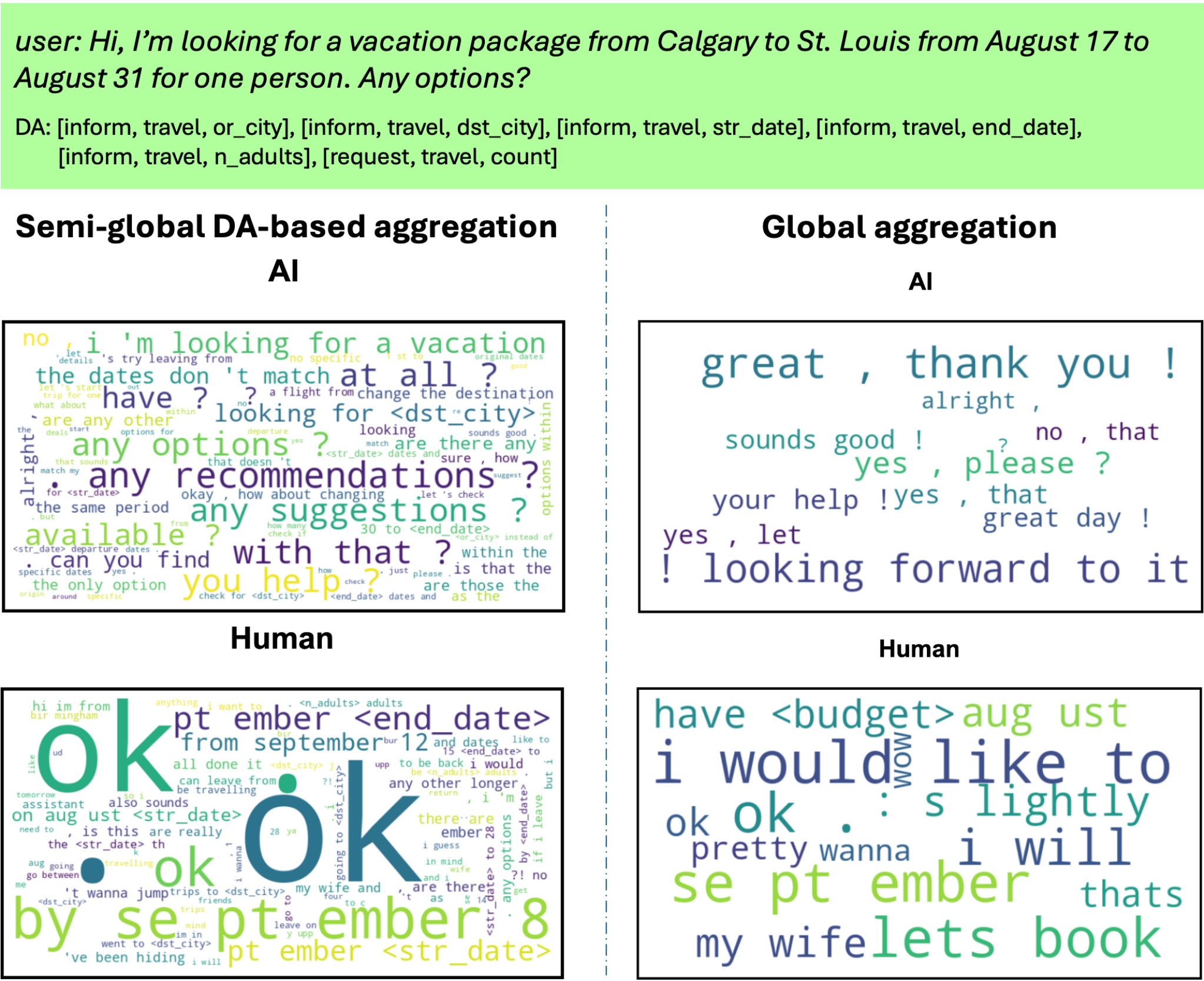}
    \caption{An example comparison of contextualized semi-global and global aggregation.}
    \label{fig:aggregation-sample}
\end{figure}

Table~\ref{tab:attribution-test} compares detection performance under different attribution methods and feature budgets per utterance. The attribution methods include Faith-SHAP~\cite{Tsai2023-faithShap}, Shapley Taylor Interaction Index~\cite{Sundararajan2020-STII}, and Integrated Gradients~\cite{Sundararajan2017-integratedGradient}, covering both white-box techniques and model-agnostic approaches that approximate the Shapley values~\cite{Shapley1953-shapley}.
Faith-SHAP consistently ranks first or second in Macro-F1, demonstrating its reliability and effectiveness in identifying salient features. Nevertheless, all attribution methods achieve strong performance under strict interpretability constraints ($>0.9$ Macro-F1), underscoring the framework's robustness across different implementations.

\subsection{Explanation Relevance: Are aggregated explanations relevant to local predictions?} 
This study applies two aggregation metrics: (1) AGG, proposed for global aggregation of local explanations~\cite{Mor2024-aggregation}, and (2) log odds ratio with informative Dirichlet prior (LOR), designed to identify disproportionate word usage between two corpora~\cite{Monroe2008-LOR}. Based on feature attributions in the training dataset, we define two corpora: the AI corpus, containing features with positive attribution toward AI predictions, and the Human corpus, containing features with negative attribution. For AGG, originally applied to ``anchors" selected as important features for a prediction of class $c$, we analogously define the anchor frequency of a token $t$ for class $c$ as its frequency in the corresponding corpus of the class $c$. LOR uses the two corpora directly as input.

Figure~\ref{fig:aopc} presents the results of the area over the perturbation curve (AOPC), of an EMMM turn-level utterance-based detection model. Aggregating local explanations based on dialogue acts yields higher AOPC scores than global aggregation, with one-sided Wilcoxon signed-rank tests on paired per-sample $AOPC_k$ values across $k\leq20$ yielding $p < 0.05$ in all settings of datasets, classes, and aggregation metrics. In particular, AOPC scores are lower for human-predicted samples, which aligns with prior studies showing that human language exhibits greater linguistic variability~\cite{MunozOrtiz2024-MGTPattern}. This variability reduces the overlap between the top 20 aggregated tokens and those in individual samples. Overall, these results indicate that our contextualization technique based on DA offers semi-global model insights that are more relevant and informative for understanding local predictions.

For qualitative analysis, Figure~\ref{fig:aggregation-sample} illustrates an example of DA-based and global aggregation. The left word cloud shows the top 20 phrases per DA, while the right shows the top 20 global phrases. The phrases are ranked by frequency-weighted LOR, highlighting high-frequency phrases disproportionally contributing to a class. Text size is scaled by both the weighted score and phrase length for clearer visualization. DA-based aggregation reveals phrases more relevant to the current utterance when comparing AI and human classifications. 
For example, AI-contributing features include question-oriented phrases such as “any options?” and “any recommendations?” when stating travel needs, whereas human-contributing features tend to convey needs directly without posing questions, and may include personal context like ``my wife", ``friends", etc. In contrast, global aggregation highlights the most influential class-specific phrases across the dataset but lacks specificity to the target utterance.
While long-form texts can be grouped by topic, task-oriented dialogues benefit from grouping by dialogue acts to capture relevant and detailed contextual explanations.

\subsection{Interpretability: Do humans prefer EMMM’s explanations?} 

\begin{figure}[t]
    \centering
    \includegraphics[width=\linewidth]{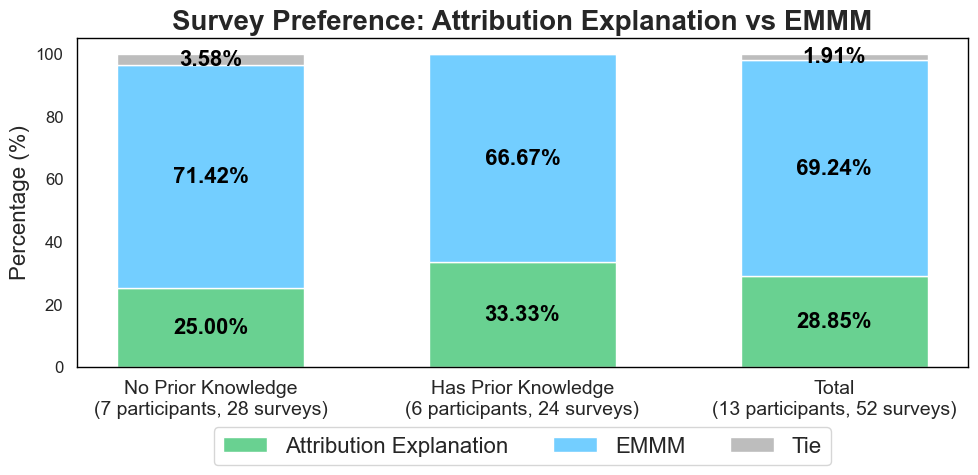}
    \caption{Human survey result on user interpretability preference between attribution explanation and EMMM. Participants are divided into two groups depending on their prior understanding of how AI models make decisions.}
    \label{fig:survey-result}
\end{figure}

We conducted a human survey to evaluate the interpretability of our output explanations. In this experiment, we randomly selected four representative samples, each covering a different dataset and prediction class. All 13 participants were asked to review these samples, resulting in a total of 52 survey responses. For each survey, participants indicate their preference between our framework’s generated explanation and a baseline attribution-based explanation. A sample of our explanation is shown in Figure~\ref{fig:demo-explanation}. The baseline attribution-based explanation displays the top 10 most important tokens and 3 dialogue acts (DAs) per utterance, sorted by descending absolute attribution scores. Attribution scores are provided for both tokens and DAs. 
Further details about the survey design are provided in Appendix~\ref{appendix:survey}.
To assess how well our explanations support both expert and non-expert users, we divided participants into two groups: one with prior knowledge of how AI models typically make decisions, and another without significant scientific training.

Figure~\ref{fig:survey-result} summarizes the survey results. In general, 69\% of the participant responses preferred the explanations generated by our framework over attribution-based methods. Participants without prior knowledge of how AI models make decisions showed an even stronger preference for our framework (71.42\%), compared to those with relevant background knowledge (66.67\%). 
These results suggest that our explanation framework offers general advantages across user groups, particularly in enhancing accessibility and interpretability for non-technical users. This highlights its potential for real-world applications where explanations need to be interpretable by a broad audience. To further support our conclusion, we evaluate the explanations generated by EMMM based on three HCI principles, with the score ratios provided in Appendix~\ref{appendix:survey}.

\subsection{Time Complexity: Can it run in real time?}
We evaluated the time complexity of our detection explanation framework by averaging the runtime over 100 randomly selected utterances from each dataset. Table~\ref{tab:time-complexity} provides a detailed breakdown of the time consumed by each step within the framework.

Report generation is highly efficient, completing within 0.5 second by leveraging pre-computed aggregation scores and natural language templates. The overall explanation process, including attribution and report generation, remains under one second, demonstrating suitability for near real-time applications. Notably, dialogue act extraction is the primary bottleneck, especially for the Frames dataset where LLM-based extraction takes over 3 seconds. This step can be replaced by a supervised model for greater efficiency, as shown by SPADE with runtime under one second. 

\begin{table}[t]
    \centering
    \begin{tabular}{llcc}
    \toprule
        \textbf{Step} & \textbf{Module} & \textbf{SPADE} & \textbf{Frames}\\
        \midrule
        2 & DA Extraction & 0.8449 & 3.4771 \\
        3 & Turn-level Detection & 0.0336  & 0.0335 \\
        4 & Attribution Explanation & 0.5157 & 0.4334 \\
        5,6 & Dialogue-level Detection & 0.3599 & 0.3633 \\
        7 & Explanation Report & 0.4379 & 0.3893 \\
        \hline
        4,7 & Explanation & 0.9536 & 0.8227 \\
        1-7 & Framework & 2.1921 & 4.6966 \\
    \bottomrule
    \end{tabular}
    \caption{Breakdown of framework average time complexity per utterance in seconds. The steps align with Figure~\ref{fig:framework}.}
    \label{tab:time-complexity}
\end{table}

\section{Conclusion} \label{conclusion}
In this paper, we present EMMM, the first explainable chatbot detection framework tailored for LLM-generated content in conversational settings. EMMM addresses critical challenges in chatbot detection, including dialogue-specific structure and the interpretability gap for non-expert users. Through a dialogue-aware architecture and an efficient selector–predictor pipeline, EMMM achieves state-of-the-art detection accuracy while delivering turn and dialogue-level explanations within real-time constraints. Our interpretability evaluation demonstrates that EMMM’s natural language explanations and semi-global visualizations significantly improve user comprehension, with a 69\% preference over the baseline approach. These contributions mark an important step toward practical, explainable, and deployable MGT detection systems for high-stakes domains such as emergency and customer service platforms.

\section{Acknowledgment}
This work was supported in part by the Australian Research Council Centre of Excellence for Automated Decision-Making and Society, and by the Australian Internet Observatory, which is co-funded by the Australian Research Data Commons (ARDC) through the HASS and Indigenous Research Data Commons.

\bibliography{main}

\clearpage 
\appendix
\section{Contextualized Semi-global Aggregation}\label{appendix:DA-aggregation}

\subsection{Text-DA Matching}
The function \textit{ExtractFeaturesPerDA(D, DA)} in Algorithm~\ref{alg:offline-aggregation} extracts all token or phrasal features in all user utterances in the dataset \textit{D} that are matched to a specific \textit{DA}.

Given an utterance ($T_{utt}$) and the associated DAs ($DA_{utt}$ with $|DA_{utt}|\geq 1$), we aim to find utterance text spans associated with each DA. An embedding-based approach is used to find matching pairs of utterance tokens and DAs:
\begin{enumerate}
    \item Convert each $DA_i$ \textit{(intent, domain, slot, value)} to a text string $T_{DA,i}$ in format ``\textit{\{intent or intent description\} \{slot and optionally (slot description)\} \{value\}}". For example, \textit{(inform, travel, or\_city, Gotham City)} is converted to ``\textit{inform or\_city (Origin city) Gotham City}". Whereas non-natural language intents like \textit{``nobook"} is replaced by their description of ``\textit{booking is failed}".
    \item Encoder inference on entire utterance ($T_{utt}$), and on each individual DA text ($T_{DA,i}\in DA_{utt}$). This study uses a paraphrase-MiniLM-L6-v2 model~\cite{Reimers2019-sentenceBert} fine-tuned for 3 epochs on a balanced dataset of positive and negative samples using CosineSimilarityLoss. For each utterance in the training set, positive samples have it paired with their associated DAs (converted to text strings and concatenated), whereas negative samples have it paired with concatenated DAs of a randomly sampled utterance.
    \item Extract token embeddings for each utterance token ($t_{utt} \in T_{utt}$), and for each DA token ($t_{DA} \in T_{DA,i}$)
    \item Calculate cosine similarity for each pair of utterance and DA tokens: $\{cos\_sim(t_{utt},t_{DA})|t_{utt} \in T_{utt}, t_{DA} \in T_{DA,i}, T_{DA,i} \in DA_{utt}\}$
    \item Similarity between an utterance token and an entire DA is defined as the maximum similarity score between the utterance token with each DA token within the DA: $s(t_{utt},T_{DA,i}) = max\{cos\_sim(t_{utt},t_{DA}) | t_{DA}\in T_{DA,i}\}$
    \item Cross-Domain Similarity Local Scaling (CSLS)~\cite{Lample2018-csls} adjusts similarity scores by increasing those for features with few close neighbors and decreasing those for features with many. This reduces the influence of tokens that are broadly similar to many DAs, or vice versa, ensuring more relevant matches are prioritized. The CSLS-adjusted similarity is computed as: $s_{csls}(t_{utt},T_{DA,i}) = 2\cdot s(t_{utt},T_{DA,i}) - r(t_{utt}) - r(T_{DA,i})$, where $r(t_{utt})$ is the average of top $k=5$ similarity scores between $t_{utt}$ and $T_{DA,i} \in DA_{utt}$, with $r(T_{DA,i})$ computed analogously.
    \item For each utterance token $t_{utt}$, match it with any DA $T_{DA,i}$ that satisfies the condition: $s_{csls}(t_{utt},T_{DA,i}) \geq min_{t_{utt}} + 0.9 \times (max_{t_{utt}} - min_{t_{utt}})$, where $min_{t_{utt}}$ and $max_{t_{utt}}$ denote the minimum and maximum $s_{csls}$ the token $t_{utt}$ has across all $T_{DA,i} \in DA_{utt}$.
\end{enumerate}

If a DA is matched to continuous text spans, n-gram phrases can be extracted. To enable investigation of phrase structure rather than specific values such as dates or locations, these texts are replaced with special tokens (e.g., \textless str\_date\textgreater) whenever they exactly match the value component of the DAs.

\subsection{Word Cloud Display}
The top phrases selected based on aggregation scores may overlap and hinder explanation interpretation. For example, ``looking forward to it!", ``looking forward", ``forward" may all be returned as important features. For clearer visualization, the word cloud displays features retained upon the following de-duplication procedure:
\begin{enumerate}
    \item \textbf{Filter duplicates}: retain phrases that either (1) do not have any other phrases containing it or (2) do not contain any other phrases. For example, ``looking forward to it!" and ``forward" would be returned, whereas ``looking forward" would be removed to reduce duplications.
    \item \textbf{Merge phrase}: merge any two phrases if there exists a consecutive overlap from the 2 ends for a minimum of $K=2$ tokens. For example, ``I'm looking forward to" would be merged with ``looking forward to it!" as ``I'm looking forward to it!" to further reduce replications.
\end{enumerate}

\section{Experimental Details}\label{appendix:experiment-details}
This section details the model configurations and training specifics for both our EMMM framework and the baseline models. All experiments were conducted on a 80GB NVIDIA A100 GPU. Whenever applicable, randomness is controlled using a seed of 2025.
 
\subsection{EMMM Framework Implementation}
This section outlines the procedure of configuration setting for the framework implementation, and examines the robustness of framework performance upon different configurations. Detection performance reported is based on the validation set. Table~\ref{tab:implementation} summarises the final framework implementation alongside the alternative options examined.

\begin{table}[ht]
    \centering
    \begin{tabular}{lc}
        \toprule
        \textbf{Module} & \textbf{Alternative Implementations} \\
        \midrule
        PLM  & \textbf{distilgpt2}, distilroberta-base, roberta-base\\
        Fusion & \textbf{average}, concatenate, max\\
        Attribution & \textbf{Faith-SHAP}, STII, Integrated Gradient\\
        \bottomrule
    \end{tabular}
    \caption{Chosen framework implementation is \textbf{bolded}.}
    \label{tab:implementation}
\end{table}

\subsubsection{PLM}
Following prior work~\cite{Schoenegger2024-XAIGT, Bafna2024-AIGTroberta}, PLMs are fine-tuned to serve as detection models during both turn-level and dialogue-level detection. Three PLMs are assessed based on their turn-level detection performance, as reported in Table~\ref{tab:PLM}. 

Overall, distilgpt2 and distilroberta-base demonstrate relatively high and stable performance across different datasets and tasks, with distilgpt2 slightly outperforming in three out of four settings. Roberta-base tends to have more unstable convergence. Distilgpt2 is selected as the base PLM in framework implementation.

\begin{table}[ht]
    \centering
    \begin{tabular}{lcc}
    \hline
        \textbf{PLM} & \textbf{DA} & \textbf{Utterance} \\
        \hline
        \textit{SPADE} & & \\
        \hline
        distilgpt2 & \underline{0.6350 (0.0077)} & \textbf{0.9181 (0.0000)} \\
        distilroberta-base & \textbf{0.6440 (0.0035)} & \underline{0.9141 (0.0000)} \\
        roberta-base & 0.5465 (0.1229) & 0.7775 (0.0000) \\
        \hline
        \textit{Frames} & & \\
        \hline
        distilgpt2 & \textbf{0.7372 (0.0067)} & \underline{0.9789 (0.0000)} \\
        distilroberta-base & \underline{0.7248 (0.0038)} & 0.9781 (0.0000) \\
        roberta-base & 0.5635 (0.1125) & \textbf{0.9802 (0.0000)} \\
    \hline
    \end{tabular}
    \caption{Comparison of Macro-F1 scores using different PLM for turn-level DA-based and utterance-based detection. Standard deviations reported in brackets. The best performance per task is \textbf{bolded}, and the second-best is \underline{underlined}.}
    \label{tab:PLM}
\end{table}

\subsubsection{Fusion}
\begin{figure}[ht]
    \centering
    \includegraphics[width=\linewidth]{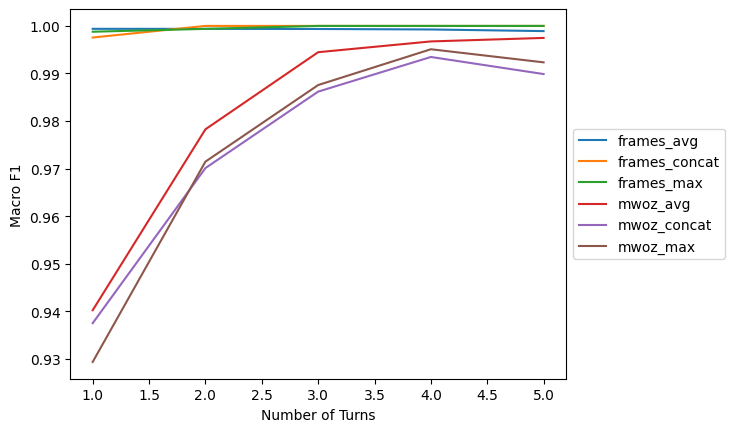}
    \caption{Comparison of Macro-F1 scores using different fusion methods for online detection when different number of turns are available progressively.}
    \label{fig:fusion}
\end{figure}

Fusion techniques are compared by measuring model's online detection performance using all DA and token features across available turns. From Figure~\ref{fig:fusion}, different fusion techniques exhibit similar performance, indicating that the framework is relatively robust to the fusion method used. Average fusion shows slightly better performance than concatenation and max fusion, and is therefore used in the framework implementation.

\subsubsection{Feature attribution}
The three feature attribution methods compared, along with their configurations, are summarized below:
\begin{itemize}
    \item Faith-SHAP~\cite{Tsai2023-faithShap}: 200 perturbations per sample.
    \item Shapley Taylor Interaction Index (STII)~\cite{Sundararajan2020-STII}: 50 perturbations per sample, with at least 1 perturbation per feature. The remaining perturbations are distributed equally across features, rounding up as necessary to ensure equal allocation.
    \item Integrated gradient~\cite{Sundararajan2017-integratedGradient}: 100 integration steps.
\end{itemize}

To select a feature attribution method, offline dialogue-level detection performance is evaluated using varying numbers of token and DA features per utterance, selected based on the rankings of their absolute attribution scores. As shown in Table~\ref{tab:attribution}, Faith-SHAP consistently ranks first or second, and is thus adopted in the framework implementation.

\begin{table}[ht]
    \centering
    \begin{tabular}{lcc}
    \hline
        \textbf{Explanation} & \textbf{1DA + 1token} & \textbf{3DA + 3token} \\
        \hline
        \textit{SPADE} & & \\
        \hline
        Faith-SHAP & \textbf{0.9470 (0.0118)} & \underline{0.9619 (0.0039)} \\
        STII & 0.8871 (0.0045) & 0.9334 (0.0135) \\
        Integrated gradient & \underline{0.9006 (0.0106)} & \textbf{0.9742 (0.0024)} \\
        
        \hline
        \textit{Frames} & & \\
        \hline
        Faith-SHAP & \underline{0.9890 (0.0012)} & \underline{0.9963 (0.0012)} \\
        STII & \textbf{0.9933 (0.0044)} & \underline{0.9963 (0.0012)} \\
        Integrated gradient & 0.9854 (0.0017) & \textbf{0.9976 (0.0000)} \\
    \hline
    \end{tabular}
    \caption{Macro-F1 scores for different explanation methods and maximum number of features per utterance. Standard deviations reported in brackets. Best scores per group are \textbf{bolded}, second-best are \underline{underlined}.}
    \label{tab:attribution}
\end{table}

\begin{table*}[t]
    \centering
    \begin{tabular}{lp{13cm}}
    \toprule
        \textbf{Model} & \textbf{Training settings and Hyperparameters}\\
        \hline
        Entropy & \begin{itemize}
                \item Tree max depth: \textbf{10},20,30,40, None
            \end{itemize}
        \\ \hline
        $\text{Raidar}_{\text{llama}}$ & 
            We directly adopt the  implementation and training details from the original work.
        \\ \hline
        Random Forest & \begin{itemize}
                \item Number of tree: 10, 50, \textbf{100}
            \end{itemize}
        \\ \hline
        MLP & 
        Models were trained using the Hyperband tuner from Keras Tuner with default settings.
        \begin{itemize}
            \item Number of hidden layers: 2,3,\textbf{4},5
            \item Number of units per layer: 16, \textbf{32}, 64
            \item Optimizer: Adam
            \item Maximum epochs: 25
            \item Early stop: True 
            \end{itemize}
        \\ \hline
        EMMM & Table~\ref{tab:implementation} outlines the framework's configurable components and their selected implementations.

        Models were trained using the Trainer class from HuggingFace Transformers with default settings unless otherwise specified below:
            \begin{itemize}
                \item Batch size: 16
                \item Epoch (turn-level DA-based detection): 15 
                \item Epoch (turn-level utterance-based detection): 10
                \item Epoch (dialogue-level detection): 5
            \end{itemize}
        \\
    \bottomrule
    \end{tabular}
    \caption{Hyperparameters and training settings for the supervised models. When applicable, final hyperparameters are bolded.}
    \label{tab:hyperparameter}
\end{table*}

\subsection{Hyperparameters and Training}
Table~\ref{tab:hyperparameter} summarizes the baseline and EMMM model hyperparameters and training details.

\section{Human Survey}\label{appendix:survey}

\begin{figure}[t]
    \centering
    \includegraphics[width=\linewidth]{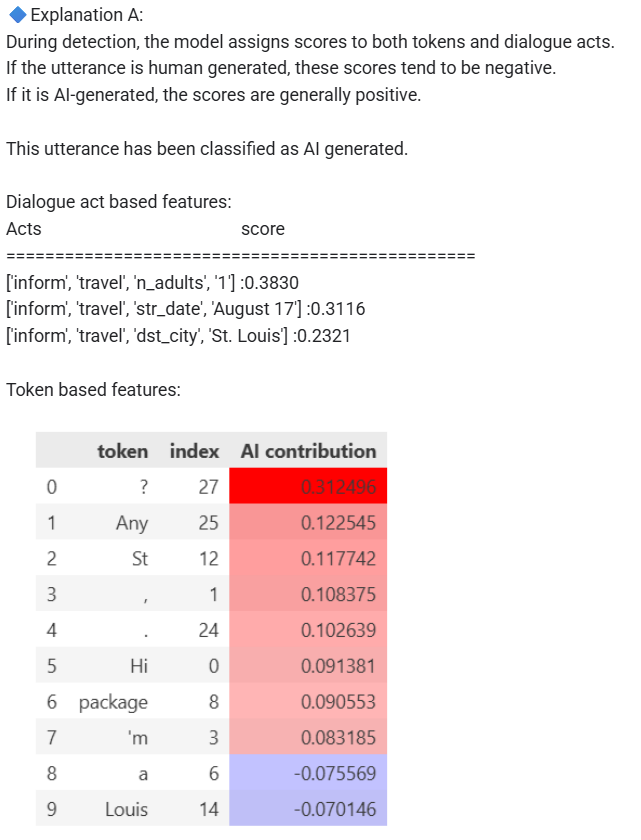}
    \caption{A demonstration of baseline attribution-based explanation.}
    \label{fig:attribution method survey}
\end{figure}

The human survey was designed to evaluate the non-expert-oriented explanations generated by our proposed framework, EMMM. Participants were asked to select their preferred explanation for a given user utterance. Figure~\ref{fig:attribution method survey} illustrates the traditional attribution-based explanations used as a baseline for comparison. Both baseline and EMMM explanations are presented with the target utterance with tokens color-coded by the attribution scores as shown at the top of Figure~\ref{fig:demo-explanation}.
To further validate our framework, we assess the alignment of EMMM’s explanations with the three HCI principles introduced in Section 3 (naturalness, flexibility, and usefulness). Participants rated the generated explanations on a 5-point preference scale (where 5 indicates perfect alignment). As shown in Table~\ref{tab:survey-hci-result}, EMMM achieves consistently high average scores across all three criteria, demonstrating strong adherence to human-centric explanation design principles. These results empirically demonstrate that our proposed method effectively delivers user-centered interpretability.

\begin{table}[t]
    \centering
    \begin{tabular}{lc}
    \toprule
        \textbf{HCI Metric} & \textbf{Average Score}\\
        \hline
        Naturalness & 4.44 \\
        Flexibility & 4.39 \\
        Usefulness & 4.39 \\
    \bottomrule
    \end{tabular}
    \caption{This table presents the scores of our framework based on three HCI principles for explanation user interface design.}
    \label{tab:survey-hci-result}
\end{table}

\section{Dataset Construction}\label{appendix:dataset-construction}
We applied the End-to-End Conversation Framework~\cite{Li2025-SPADE} using Qwen2.5-32B~\cite{qwen2.5} to generate synthetic dialogues based on bona fide samples from the Frames dataset~\cite{Schulz2017-Frame}. Table~\ref{tab:frames-statistics} provides an overview of the statistics of the synthetic dataset generated. 

The End-to-End Conversation Framework involves two LLMs that simulate a dialogue by taking on the roles of user and system to collaboratively pursue the user’s goal~\cite{Li2025-SPADE}. Adaptations were required to address the increased goal complexity in the Frames dataset, in which users received alternative goals after failing to complete the previous ones or were asked to terminate the conversation~\cite{Schulz2017-Frame}. To avoid the system repeatedly confirming successful searches and resulting in overly short dialogues, we guide dialogue progression by supplying the system with both user goals and their outcomes. The goal template used aligns with the description of original Frames dataset construction~\cite{Schulz2017-Frame}. Table~\ref{tab:prompt-goal} presents the prompt structure used to extract goals and outcomes from bona-fide dialogues. To ensure the system does not reference unrevealed user goals, we introduce an admin LLM that monitors the current goal based on dialogue history, mirroring the dynamic goal updates mechanism in the original Frames data collection process.

At each turn:
\begin{itemize}
    \item \textbf{User}: generates an utterance provided with all initial and alternative goals. Prompt structure is shown in Table~\ref{tab:prompt-user}.
    \item \textbf{Admin}: determines the user’s current goal from the chat history. Prompt structure is shown in Table~\ref{tab:prompt-admin}.
    \item \textbf{System}: generates an utterance based on all goals up to the current one, along with their outcomes. Prompt structure is shown in Table~\ref{tab:prompt-system}.
\end{itemize}

\begin{table}
  \centering
  \begin{tabular}{lc}
    \toprule
    \textbf{Metrics} & \textbf{Values}\\
    \midrule
    \#dialogues  & 1364 \\
    \#synthetic user utterances & 6082 \\
    \midrule
    Avg. \#user utterances per dialogue &  4.46 \\
    Avg. \#user words per dialogue & 64.38 \\
    Avg. \#words per user utterance & 14.44 \\
    \bottomrule
  \end{tabular}
  \caption{Statistics of the synthetic dataset created based on the Frames dataset.}
  \label{tab:frames-statistics}
\end{table}

\begin{table*}[ht]
    \centering
    \begin{tabular}{p{2cm}p{14cm}}
    \hline
        \textbf{Components} & \textbf{Prompt}\\
    \hline
    Goal & Generate the progression of goals and outcomes for the user in the dialogue. \par

    **GOAL**: Defines user's requirements for vacation packages, including origin, destination(s), dates, number of travelers, budget, flexibility, and preferences. Each goal should reflect either:
    
    1. The initial request from the user, or
    
    2. An alternative suggested by the user after the system fails to meet the previous goal.\par
    
    If a goal was unsuccessful, the user either ended the dialogue or continued with an **alternative goal**, which must begin with:  
    “If nothing matches your constraints, ...”\par
    
    Please differentiate between multiple options within the same goal and alternative goals. They are characterized as follows: 
    
    1. Options within the same goal: The user modifies previously specified constraints voluntarily to explore and compare different options, even when the system has already returned packages that match their earlier constraints.
    
    2. Alternative goal: The user modifies constraints as a fallback because the system was unable to find any matching packages with the original constraints. This must start with ``If nothing matches your constraints, ...". Alternative goals can also include multiple options within the same goal.\par
    
    Goal Templates: 
    
    For the initial goal:
    
    \textless GOAL\textgreater Find a vacation between [START DATE] and [END DATE] for [NUM ADULTS] adults and [NUM CHILDREN] kids. You leave from [ORIGIN CITY]. You want to go to [DESTINATION CITY]. You are travelling on a budget and would like to spend at most \$[BUDGET]. \textless /GOAL\textgreater
    
    For any subsequent goal:
    
    \textless ALT\_GOAL\textgreater If nothing matches your constraints, [describe alternative criteria change like changing dates, destinations, budget, etc.] \textless /ALT\_GOAL\textgreater

\\

    Outcome & **OUTCOME**: Defines the vacation packages or suggestions the system returned in response to each goal. Include specific package details mentioned in the dialogue (e.g., hotel names, dates, locations, cost, star ratings, amenities, etc.).
    
\\
    
    Examples &  [demonstrations]
    
    \\
    
    Response & \#\#\# Your Task:
    
    Now, extract the progression of \textless GOAL\textgreater and \textless OUTCOME\textgreater tags for the following dialogue. 
    
    Think about the goal progression using \textless THINK\textgreater and \textless /THINK\textgreater, focus on whether the user has multiple options within one goal.
    
    For the one initial goal, use \textless GOAL\textgreater ... \textless /GOAL\textgreater. 
    
    For any subsequent alternative goal, use \textless ALT\_GOAL\textgreater ... \textless /ALT\_GOAL\textgreater.
    
    Every alternative goal must begin with ``If nothing matches your constraints,"
    
    Include all relevant database information under \textless OUTCOME\textgreater.
\\
    \hline
    \end{tabular}
    \caption{Prompt structure to extract user goals and outcomes from a dialogue.}
    \label{tab:prompt-goal}
\end{table*}

\begin{table*}[ht]
    \centering
    \begin{tabular}{p{2cm}p{14cm}}
    \hline
        \textbf{Components} & \textbf{Prompt}\\
    \hline
    Instruction & Based on the conversation so far, which goals is the user currently expressing? 

    \\

    Chat History & Conversation: [chat history]

    \\
    
    Goal & Goals: [user goal]

    \\
    
    Response & Your final response should be in format of \textless goal\textgreater x \textless /goal\textgreater, where x is the index of the goal which the user is currently working on.
    \\
    \hline
    \end{tabular}
    \caption{Prompt structure for admin simulation, to identify the current user goal based on the chat history.}
    \label{tab:prompt-admin}
\end{table*}

\begin{table*}[ht]
    \centering
    \begin{tabular}{p{2cm}p{14cm}}
    \hline
        \textbf{Components} & \textbf{Prompt}\\
    \hline
    Task & Task: Simulate as an user with a particular goal and generate one response to a task oriented dialogue system. Response must start with ``user: ". After you achieved all your goals, end the conversation and generate ``[END]" token. If you think the system cannot help you or the conversation falls into an infinite loop, generate a ``[STOP]" token. The response must be one line only!\par

    The information you can ask for or provide (include everything is not mandatory): [ontology slot value]
    
    Information with ``mask\_token” specified must be replaced by corresponding token in your response. Do not ask for or provide other information. You do not need to confirm details with the system unless it is ambiguous.

\\

    Example & Here are demonstration dialogues unrelated to your own goal: [demonstrations]
    
    Do not copy anything from the demonstration!
    
\\
    
    Goal &  Here is your goal: [goal]
    
    Move through the goals in sequential order when preceding goals cannot be completed. 

\\

    Response & 
    You should end conversation only once a booking is successfully made by the system, or that none of the goals can be satisfied.
    
    Do not generate "END" when requesting a booking.
    
    Do not directly copy from the goal, be creative in generating the user response like a human.
    
    The user response must be within 20 words using natural and fluent English.

\\

    Chat History & Chat history between you and the system: [chat history]
\\
    \hline
    \end{tabular}
    \caption{Prompt structure for user simulation, to generate the next user response given a chat history.}
    \label{tab:prompt-user}
\end{table*}

\begin{table*}[ht]
    \centering
    \begin{tabular}{p{2cm}p{14cm}}
    \hline
        \textbf{Components} & \textbf{Prompt}\\
    \hline
    Task & Task: Simulate as a task oriented dialogue system and generate one response to a user. Response must start with ``system: ". If and only if the user has no more queries or generated ``[END]", end the conversation and generate ``[END]" token. If you think the conversation falls into an infinite loop, generate a ``[STOP]" token.\par

    The information you can ask for or provide (include everything is not mandatory): [ontology slot value]
    
    Information with ``mask\_token” specified must be replaced by corresponding token in your response. Not all information is mandatory, and you do not need to provide information not asked by the user, nor to confirm if they need it. Do not ask for or provide other information. Do not repeat yourself unless asked by the user. You do not need to confirm details with user unless it is ambiguous.

\\

    Example & Here are demonstration dialogues: [demonstrations]
    
    Do not copy anything from the demonstration!
    
\\
    
    Goal &  Here are the user goals and the outcomes of searching for relevant vacation packages: [user goals and outcomes]

\\

    Response & 
    Before making suggestions or bookings, check whether the user has specified preference or flexibility on all critical information: location (dst\_city, or\_city), time (str\_date, end\_date, duration), number of people (n\_adults, n\_children), and budget.
    
    Identify any missing critical information based on the chat history: ``I need to confirm: \textless at most 2 missing items, or None if all are provided\textgreater"
    
    Then, generate your booking assistant response starting with: ``system: "
    
    In the booking assistant response, do not directly copy from the goal or outcome, do not say ``I need to confirm", be creative and respond like a human.
    
    The booking assistant response must be within 20 words using natural and fluent English.

\\

    Chat History & Chat history between you and the user: [chat history]
\\
    \hline
    \end{tabular}
    \caption{Prompt structure for system simulation, to generate the next system response given a chat history.}
    \label{tab:prompt-system}
\end{table*}

\end{document}